\documentclass[letterpaper]{article} 
\usepackage{aaai2026}  

\usepackage{times}  
\usepackage{helvet}  
\usepackage{courier}  
\usepackage[hyphens]{url}  
\usepackage{graphicx} 
\usepackage{natbib}  
\usepackage{caption} 
\usepackage{algorithm}
\usepackage{algorithmic}

\usepackage{multirow} 
\usepackage{array}
\usepackage{booktabs}
\usepackage{amssymb} 

\usepackage{amsmath}
\usepackage{amsfonts}
\usepackage{dsfont}
\usepackage{mathrsfs}
\usepackage{subcaption}

\usepackage{newfloat}
\usepackage{listings}
\DeclareCaptionStyle{ruled}{labelfont=normalfont,labelsep=colon,strut=off} 
\floatstyle{ruled}
\newfloat{listing}{tb}{lst}{}
\floatname{listing}{Listing}
\title{Query-aware Hub Prototype Learning \\ for Few-Shot 3D Point Cloud Semantic Segmentation}
\author{
    Yilin Zhou\textsuperscript{\rm 1}\equalcontrib, 
    Lili Wei\textsuperscript{\rm 1}\equalcontrib, 
    Zheming Xun\textsuperscript{\rm 1}\equalcontrib, 
    Congyan Lang\textsuperscript{\rm 1}\thanks{Corresponding author.}, 
    Ziyi Chen\textsuperscript{\rm 1}
}
\affiliations{
    \textsuperscript{\rm 1}Key Laboratory of Big Data \& Artificial Intelligence in Transportation (Ministry of Education), School of Computer and Information Technology, Beijing Jiaotong University
}

\begin{document}

\maketitle

\begin{abstract}
Few-shot 3D point cloud semantic segmentation (FS-3DSeg) aims to segment novel classes with only a few labeled samples. However, existing metric-based prototype learning methods generate prototypes solely from the support set, without considering their relevance to query data. This often results in prototype bias, where prototypes overfit support-specific characteristics and fail to generalize to the query distribution, especially in the presence of distribution shifts, which leads to degraded segmentation performance. To address this issue, we propose a novel Query-aware Hub Prototype (QHP) learning method that explicitly models semantic correlations between support and query sets. Specifically, we propose a Hub Prototype Generation (HPG) module that constructs a bipartite graph connecting query and support points, identifies frequently linked support hubs, and generates query-relevant prototypes that better capture cross-set semantics. To further mitigate the influence of bad hubs and ambiguous prototypes near class boundaries, we introduce a Prototype Distribution Optimization (PDO) module, which employs a purity-reweighted contrastive loss to refine prototype representations by pulling bad hubs and outlier prototypes closer to their corresponding class centers.  Extensive experiments on S3DIS and ScanNet demonstrate that QHP achieves substantial performance gains over state-of-the-art methods, effectively narrowing the semantic gap between prototypes and query sets in FS-3DSeg.

\end{abstract}


\section{Introduction}

Point cloud semantic segmentation assigns semantic labels to each point in a 3D point cloud and is essential for applications like autonomous driving and robotics. Although fully supervised methods \cite{qi2017pointnet++, Lin_2020_CVPR, qian2022pointnext} have made significant progress, they rely heavily on costly manual annotations and struggle to generalize to novel classes. To address these challenges, few-shot 3D point cloud segmentation (FS-3DSeg) has gained increasing attention, aiming to learn generalizable models from abundant base class data and adapt the model to novel classes with only a few labeled point clouds.

\begin{figure}[t]
\centering
\includegraphics[width=0.9\columnwidth]{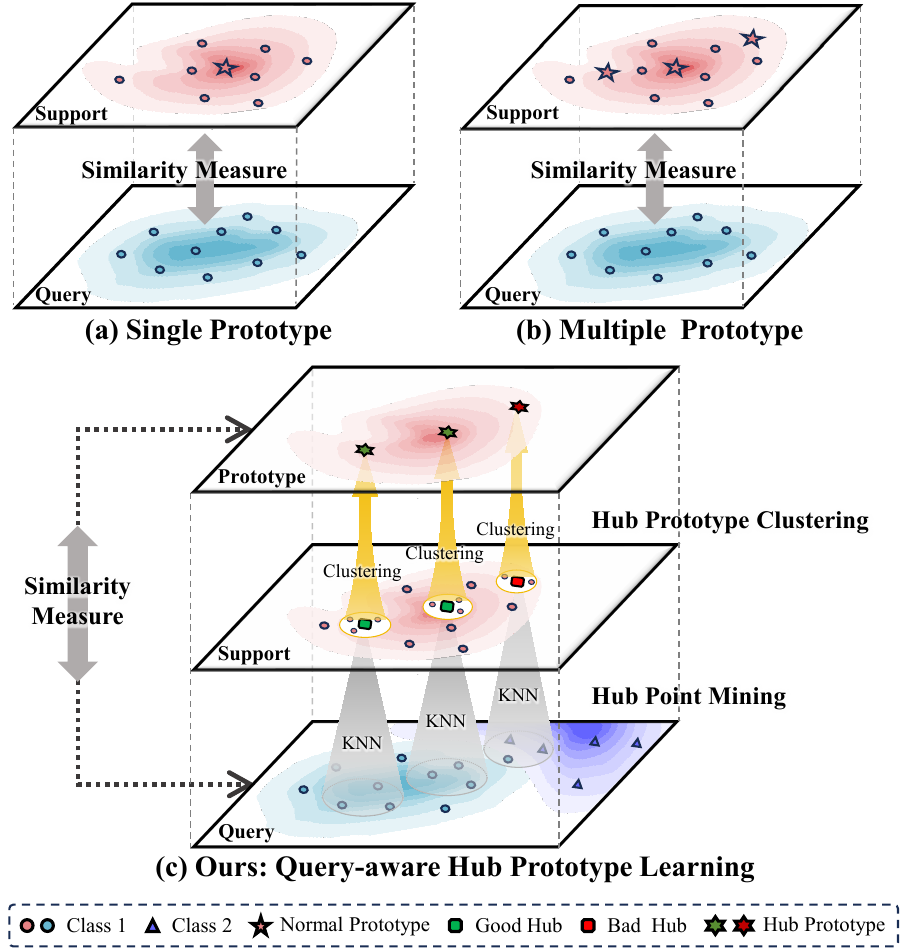} 
\caption{
Few-shot 3D point cloud semantic segmentation approaches.
(a)(b) Previous prototype learning methods generate prototypes solely based on support points.
(c) We propose a Query-aware Hub Prototype Learning method that generates prototypes more closely related to query points.
}
\label{fig1}
\end{figure}

Recent FS-3DSeg methods typically adopt metric-based prototype learning frameworks, where prototypes are derived from a few labeled support point clouds, and the unlabeled query set is segmented by measuring similarity between query points and these prototypes. As illustrated in Figure \ref{fig1} (a)(b), these methods can be broadly categorized into two groups: single-prototype approaches \cite{3DV2022BFG, TIP2023QGPA, liu2024DPA}, which generate global class prototypes via masked average pooling, and multi-prototype methods \cite{CVPR2021MPTI,an2024COSeg}, which enhance prototype diversity through strategies like Farthest Point Sampling (FPS) and local clustering.
Ideally, prototypes should serve as semantic bridges between support and query sets, requiring strong alignment with query semantics. However, existing methods generate prototypes solely from the support set, emphasizing internal representativeness or diversity while ignoring semantic relevance to the query. This often causes prototype bias, especially under distribution shifts between support and query sets. For example, intra-class variations (e.g., square vs. round tables) may share limited similarity, causing prototypes to overfit support-specific traits and poorly represent diverse queries.
Moreover, uniform sampling strategies like FPS often introduce redundant or query-irrelevant prototypes, further compromising segmentation accuracy. To address these challenges, it is essential to develop a query-aware prototype generation mechanism to narrow the semantic gap between prototypes and the query set and improve segmentation performance.

Realizing the above issues, we propose a \textbf{Q}uery-aware \textbf{H}ub \textbf{P}rototype (\textbf{QHP}) learning method, as depicted in Figure \ref{fig1} (c).
Hubs~\cite{radovanovic2009nearest} refer to data points that frequently appear among the nearest neighbors of many other points. Therefore, hubs naturally reflect support-query semantic correlation and are well-suited as prototypes. 
In recent few-shot image classification studies, some methods~\cite{Fei_2021_ICCV, trosten2023hubs, TANG2025130879}
regard hubs as a nightmare and seek to avoid them, worrying that when a support point is a hub, many query points may be retrieved regardless of their true classes.
However, we argue that:
\textbf{(1) Not all hubs are harmful.} Hubs that emerge within the same class (\textit{i.e.}, good hubs) can capture accurate support-query relationships and serve as effective query-aware prototypes, which helps mitigate prototype bias.
\textbf{(2) The harmful impact of bad hubs is limited in FS-3DSeg.} Since each support sample contains numerous points and provides richer point-level supervision, each query point’s segmentation can be determined by multiple support-query matches, reducing the risk of being misled by any single bad hub.
Therefore, instead of suppressing hubs, we leverage them to bridge the semantic gap between support and query, and propose to learn query-aware hub prototypes. Notably, to further mitigate the influence of bad hubs,
we optimize their distributions by pulling those near class boundaries closer to corresponding class centers.

The proposed QHP approach
introduces two key components: the \textbf{H}ub \textbf{P}rototype \textbf{G}eneration (\textbf{HPG}) module and the \textbf{P}rototype \textbf{D}istribution \textbf{O}ptimization (\textbf{PDO}) module.
Specifically, HPG explicitly models semantic correlations between support and query sets to learn query-relevant hub prototypes. It constructs a bipartite graph connecting query and support points, identifies support hubs with high linking frequency, and performs local clustering around each hub to generate query-relevant prototypes that better capture cross-set semantics. Query segmentation can be conducted by measuring similarities between query points and these hub prototypes.
To further mitigate the influence of bad hubs and ambiguous prototypes near class boundaries, we propose a PDO module during training. PDO constructs a global association graph to identify bad hubs, and adopts a purity-reweighted contrastive loss to pull bad hubs and outlier prototypes toward their corresponding class centers.
By jointly leveraging the HPG and PDO modules, our QHP facilitates more query-relevant and discriminative prototype learning, effectively narrowing the semantic gap between prototypes and query sets and yielding improved performance in the FS-3DSeg task.

Our main contributions can be summarized as follows:

\begin{itemize}
    \item We propose a novel Query-aware Hub Prototype (QHP) Learning method for FS-3DSeg, which explicitly models semantic correlations between support and query sets to generate query-relevant prototypes,    addressing prototype bias and narrowing the semantic gap.     
    \item We propose a Hub Prototype Generation (HPG) module to identify support hubs and generate query-relevant hub prototypes that better capture cross-set semantics.

    \item We design a Prototype Distribution Optimization (PDO) module, optimizing the distributions of bad hubs and outlier prototypes via a purity-reweighted contrastive loss.

    \item Extensive experiments on S3DIS and ScanNet demonstrate that QHP achieves state-of-the-art performance.

\end{itemize}

\begin{figure*}[t]
\centering
\includegraphics[width=0.98\textwidth]{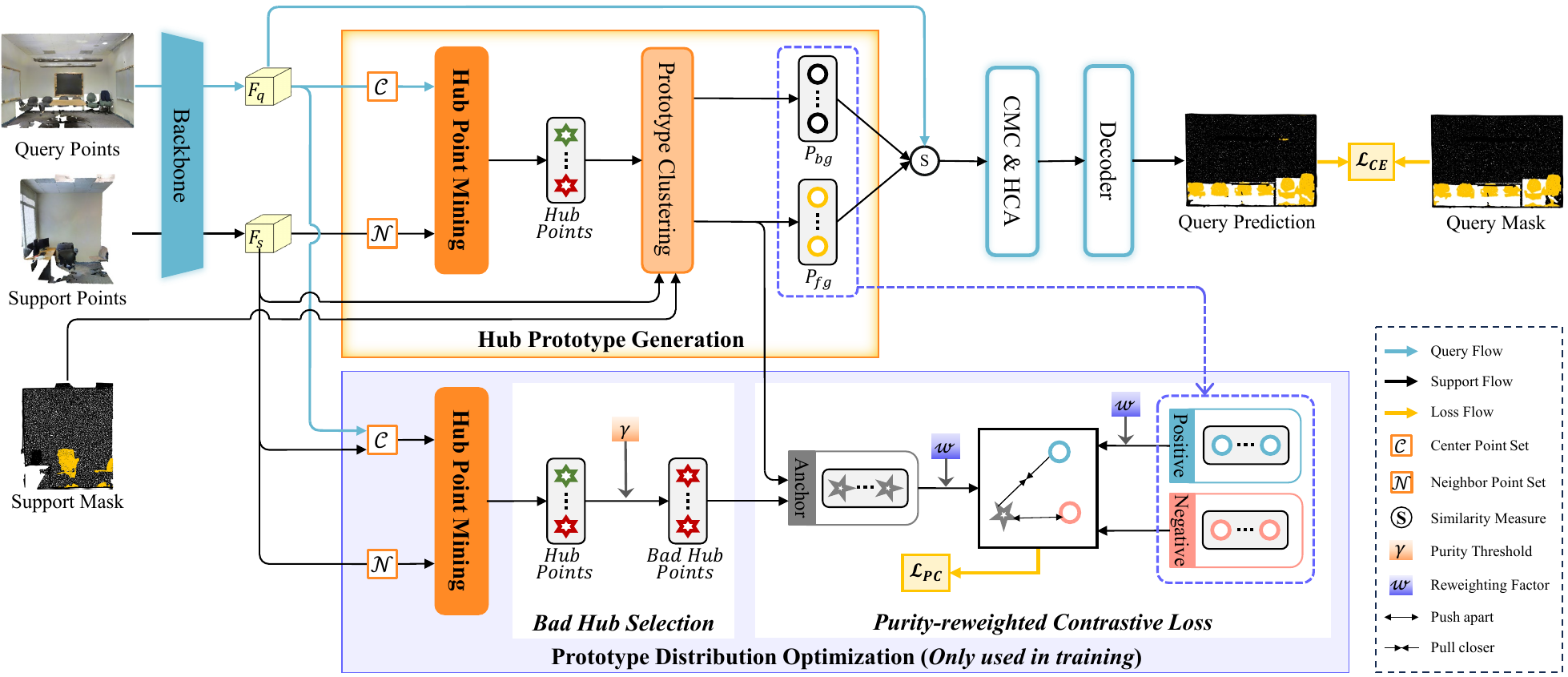}
\caption{The framework of our Query-aware Hub Prototype Learning method. 
Initially, we design an HPG module to select support hubs and generate query-relevant hub prototypes.
Moreover, during training, a PDO module is integrated to optimize the distribution of bad hubs and outlier prototypes.
For clarity, we present the model under the 1-way 1-shot setting.}
\label{fig2}
\end{figure*}

\section{Related Work}
\subsection{3D Point Cloud Semantic Segmentation}


Recent 3D point cloud segmentation methods can be broadly categorized into MLP-based (\textit{e.g.}, PointNet~\cite{pointnet} and RandLA-Net~\cite{RandLA-Net}), convolution-based (\textit{e.g.}, PointCNN~\cite{PointCNN}, KPConv~\cite{KPConv}, and RandLA-Net~\cite{RandLA-Net}), and Transformer-based approaches such as Point Transformer~\cite{PointTransformer} and Point Transformer V2~\cite{PointTransformerV2}. Although these methods demonstrate strong performance through local feature aggregation or self-attention mechanisms, they typically require expensive, large-scale annotations and struggle to generalize to novel classes unseen during training.

\subsection{Few-shot 3D Point Cloud Segmentation}
Recent FS-3DSeg methods primarily adopt the prototype-based paradigm built upon metric learning. These methods can be broadly categorized into single-prototype and multi-prototype approaches.
Single-prototype methods summarize each class using a single representative prototype from the support set. For instance, ProtoNet defines the prototype as the class-wise mean of support features. To mitigate distribution shifts between support and query sets, 2CBR~\cite{zhu20232CBR} explicitly models such biases, and DPA~\cite{liu2024DPA} employs query-guided attention to generate task-adaptive prototypes. However, these methods lack prototype diversity and are unsuitable for handling complex data.
To capture intra-class variations, multi-prototype approaches generate multiple prototypes per class. AttMPTI~\cite{CVPR2021MPTI} employs farthest point sampling (FPS) to extract diverse local prototypes. Stratified Transformer~\cite{lai2022stratified} combines hierarchical sampling with cross-window self-attention. COSeg~\cite{an2024COSeg} maintains a momentum-updated pool of base class prototypes.
Despite these advances, most methods rely solely on support data to generate prototypes, yielding prototypes biased toward support distribution and poorly aligned with queries, thus limiting generalization to novel classes.

\subsection{The Hubness Phenomenon and Hubs}
Hubness~\cite{radovanovic2010hubs,radovanovic2009nearest} describes the tendency of certain points, called \textit{hubs}, to appear frequently in nearest-neighbor lists. It has been studied in areas like multi-view clustering~\cite{Xu_2025_CVPR} and cross-modal retrieval~\cite{bogolin2022cross, Retrieval_hub_EMNLP}. In few-/zero-shot classification tasks, prior works~\cite{DinuB14, xiao2021one, CheraghianRCP19, trosten2023hubs} mostly view hubs as harmful, as query points may be misclassified when dominated by support hubs from different classes. In contrast, we argue that good hubs are beneficial and are primary in our scenario. We thus exploit hubs via query-aware hub prototype learning and mitigate bad hub distance optimization to narrow query-support gaps.

\begin{figure*}[t]
\centering
\includegraphics[width=1.0\textwidth]{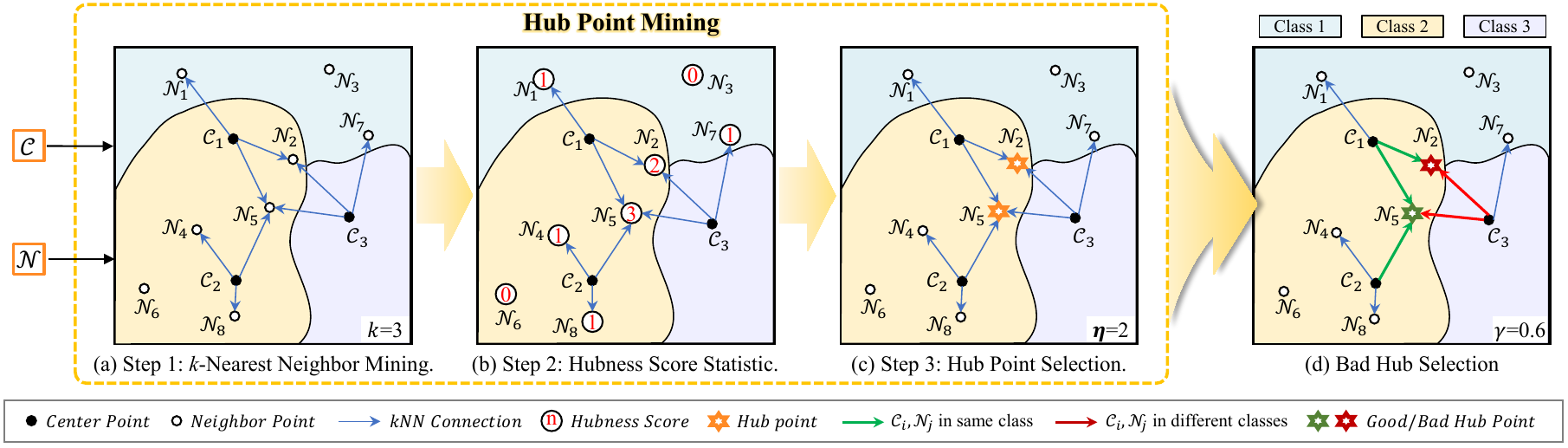} 

\caption{Illustration of Hub Point Mining and Bad Hub Selection modules. We give an example under hyperparameters $k=3$, $\eta=2$, and $\gamma=0.6$. (a)--(c) Hub Point Mining: initially, using center points $\{\mathcal{C}_1, \mathcal{C}_2, \mathcal{C}_3\}$ and neighbor points $\{\mathcal{N}_1, \dots, \mathcal{N}_8\}$ as input, a $k$NN graph is constructed with $k=3$. After calculating hubness scores, the top $\eta=2$ points with the highest hubness scores are selected as hubs. (d) Bad Hub Selection: Hubs with purity below the threshold $\gamma=0.6$ are selected as bad hubs.
}
\label{fig3}
\end{figure*}

\section{Method}

\subsection{Problem Formulation and Overview}

\textbf{Problem Formulation.} FS-3DSeg aims to predict per-point semantic labels for query point clouds using a few labeled support samples. Episodic learning \cite{CVPR2021MPTI} is typically employed to simulate the few-shot learning process, where each \(C\)-way \(K\)-shot episode includes a support set \(S = \{(I_s^{c,k}, M_s^{c,k})_{k=1}^{K}\}_{c=1}^{C}\) and a query set \(Q = \{(I_q^l, M_q^l)\}_{l=1}^{L}\). Each point cloud \(I_s^{c,k}, I_q^l \in \mathbb{R}^{T \times (3 + f_0)}\) contains \(T\) points, each represented by 3D coordinates and auxiliary features (\textit{e.g.}, RGB). \(M_s^{c,k} \in \{0,1\}^T\) denotes the binary ground truth (GT) mask indicating whether each point in \(I_s^{c,k}\) belongs to class \(c\), while \(M_q^l \) denotes the GT labels for the query point cloud $I_q^l$. 
During inference, the goal is to predict the query labels $\hat{M}_q$ for the query points in $I_q$ under the guidance of the support set $S$.

\noindent \textbf{Overview.} Figure \ref{fig2} illustrates the architecture of the proposed QHP framework, comprising two key components: a Hub Prototype Generation (HPG) module and a Prototype
Distribution Optimization (PDO) module. We first use a shared backbone $\varPhi$ to extract point-wise features from the support and query point clouds: ${F}_s=\varPhi \left( I_{s} \right) \in \mathbb{R} ^{C\times K\times T\times D}$ and ${F}_q=\varPhi \left( I_{q} \right) \in \mathbb{R} ^{L\times T\times  D}$, where $D$ is the channel dimension. In the HPG module, a Hub Point Mining (HPM) module identifies hub points from $S$, which are used to generate hub prototypes \(P\) via local clustering. These prototypes are matched with query features ${F}_q$ through similarity measures, and further refined via the CMC and HCA modules \cite{an2024COSeg} to yield query predictions $\hat{M}_{q}$. To mitigate the influence of bad hubs and ambiguous prototypes, our PDO module identifies bad hubs by thresholding their purity and applies a Purity-reweighted Contrastive (PC) loss to promote intra-class compactness. 
During training, our model is jointly optimized by a cross-entropy (CE) loss and the proposed PC loss.

Subsequently, we provide a detailed description of the HPG module, PDO module and each loss as below.

\subsection{Hub Prototype Generation}
To mitigate prototype bias, we propose an HPG module. It first identifies frequently occurring support hubs via a Hub Point Mining (HPM) module, then generates query-relevant hub prototypes through Hub Prototype Clustering.

\noindent \textbf{Hub Point Mining.} HPM identifies hub points through three sequential steps, as illustrated in Figure~\ref{fig3}(a)--(c).

\textit{Step 1: $k$-Nearest Neighbor Mining.}
Given a center point set \( \mathcal{C} \) and a neighbor point set \( \mathcal{N} \), a bipartite graph is constructed by connecting each center point \( c \in \mathcal{C} \) to its \( k \)-nearest ($k$NN) neighbors in  \( \mathcal{N} \) via cosine similarity measure. The \( k \)-nearest neighbors of \( c \) are formulated as \( k\text{NN}(c,\mathcal{N}) \).

\textit{Step 2: Hubness Score Statistic.}
The hubness score $s(n)$ quantifies how frequently a point \( n \in \mathcal{N} \) is selected as neighbors by all center points in \( \mathcal{C} \), defined as:
\begin{equation}
s(n) = \sum_{c \in \mathcal{C}} \mathds{1}\left( n \in k\text{NN}(c,\mathcal{N}) \right) + \varepsilon, \label{eq:1}
\end{equation}
where \( \mathds{1}(\cdot) \) denotes the Iverson bracket indicator function, returning 1 if the condition holds and 0 otherwise. 
A small positive constant \( \varepsilon \) is added to avoid zero scores caused by potential outliers, ensuring $s(n)>0$ for all $n$.
The collective hubness scores for all points in \( \mathcal{N} \) are denoted as \( s(\mathcal{N}) \).

\textit{Step 3: Hub Point Selection.}
To identify nodes most frequently regarded as neighbors by center points, we select a subset $\mathcal{H} \subseteq \mathcal{N}$ consisting of the top-$\eta$ neighbor points with the highest hubness scores, defined as:  
\begin{equation}
\mathcal{H} = \{ n \in \mathcal{N} \mid s(n) \in \text{Top-}\eta(s(\mathcal{N})) \}. \label{eq:2}
\end{equation}

\noindent \textbf{Hub Prototype Clustering.} Before prototype clustering, we treat points in the query set \(Q\) and support set \(S\) as input node sets \(\mathcal{C}\) and \(\mathcal{N}\) of the HPM module, and select $\text{Top-}\eta$ hub points for each class to construct hub point set $\mathcal{H}$. 

A prototype set \( P=P_{fg} \cup P_{bg} \), where \( P_{fg} \) and \( P_{bg} \) denote foreground/background prototypes, is generated by conducting point-to-seed clustering \cite{CVPR2021MPTI} on support features localized around each hub point, defined as:
\begin{equation}
\begin{aligned}
P_{fg}=\mathcal{F} _{clus}\left( F_s\odot M_s, \mathcal{H}_{fg} \right) , \mathcal{H}_{fg}=\mathcal{H}\odot M_s,
\\
P_{bg}=\mathcal{F} _{clus}\left( F_s\odot \neg{M}_{s}, \mathcal{H}_{bg} \right) , \mathcal{H}_{bg}=\mathcal{H}\odot \neg{M}_{s}, \label{eq:3}
\end{aligned}
\end{equation}
where \( \odot \) denotes the Hadamard product; \( M_s \) and \( \neg{M}_s \) are the GT mask and its inverse for support set; \( \mathcal{H}_{fg} \) and \( \mathcal{H}_{bg} \) are foreground/background hub point subsets; and \( \mathcal{F}_{clus} \) denotes the clustering operation.
After that, we obtain $\eta$ prototypes per class, yielding a total of \( (C + 1) \cdot \eta \) prototypes.

Notably, although support hubs \( \mathcal{H} \) originate from \( S \), they are geometrically closer to points in \( Q \) as they retain only those support points that best match the query distribution. Consequently, the derived hub prototypes are more aligned with \( Q \) in the metric space, facilitating improved prototype-query matching and enhanced segmentation performance.



\subsection{Prototype Distribution Optimization}

In the PDO module, we select potential bad hubs, and then adopt a Purity-reweighted Contrastive (PC) loss to suppress these bad hubs and optimize the prototype distribution.

\noindent \textbf{Bad Hub Selection.} 
To select bad hubs, we construct a global association graph via $k$NN algorithm, where we merge points from $Q$ and $S$ to form the center point set, treat \( S \) as neighbor point set \( \mathcal{N} \), and then utilize the HPM module to identify hub points \( \mathcal{H} \) from support set \( S \).


After that, we focus on identifying all potential bad hubs within \( \mathcal{H} \), \textit{i.e.}, those exhibiting stronger connections to center points belonging to different classes, as illustrated in Figure~\ref{fig3}(d). This process contains three steps:

First, we compute the number of times each hub $h$ is connected to center points of the same class, denoted as $t(h)$:
\begin{equation}
t(h) = \sum_{c \in \mathcal{C}} \mathds{1}\left( h \in k\text{NN}(c, \mathcal{N}) \right) \cdot \mathds{1}\left( M_h = M_c \right), \label{eq:4}
\end{equation}
\noindent where \( M_c, M_h \) are class labels of \( c \) and \( h \), respectively.
 
Next, a purity \( \mathscr{P}(h) \) is defined to represent the proportion of connections to center points of the same class, given by:
\begin{equation}
\mathscr{P}(h) = {t(h)}/{s(h)}. \label{eq:5}
\end{equation}
 
Finally, the bad hub point set \( \mathcal{BH} \) is filtered out using a purity threshold \( \gamma \in (0,1) \), formulated as:
\begin{align}
\mathcal{BH} &= \left\{ h \in \mathcal{H} \mid 
    \mathscr{P}(h) < \gamma
\right\}. \label{eq:6}
\end{align}

\noindent \textbf{Purity-reweighted Contrastive Loss.} To further eliminate the influence of bad hubs and outlier prototypes, we aim to pull them back to their cluster centers. A typical solution is contrastive learning~\cite{Wang_2021_CVPR}, which has been widely adopted in various areas to pull positive pairs closer and push negative pairs apart. 
Despite these successes, contrastive loss has not been explored in FS-3DSeg.
However, directly applying standard contrastive loss to bad hub anchors is suboptimal: low-purity anchors, which have high similarity to samples from other classes, tend to lie far from their true class centers and are more likely to be confused, thus requiring stronger guidance to be correctly aligned.
To tackle this issue, we propose a Purity-reweighted Contrastive (PC) loss, which dynamically adjusts the attraction strength based on the purity of each anchor sample. 

We first introduce a purity reweighting factor $w(a)$ to quantify the strength with which anchor $a \in  \mathcal{A} = \{P_{fg} \cup \mathcal{BH} \}$ is pulled toward positive prototypes, formulated as:
\begin{equation}
w(a) =
\begin{cases}
1 - \mathscr{P}(a) & \text{if} \quad a \in \mathcal{BH}\\
1 & \text{otherwise} \quad a \in P_{fg} 
\end{cases}, \label{eq:8}
\end{equation}
where the first line assigns a higher weight ($w(a) \to 1$) inversely proportional to purity for bad hub anchors ($a \in \mathcal{BH}$) with low purity ($\mathscr{P}(a) \to 0$), strongly pulling them toward class centers; the second line assigns a fixed weight ($w(a) = 1$) for all foreground prototype anchors ($a \in P_{fg}$), ensuring intra-class compactness and inter-class discriminability.


Based on the defined purity reweighting factor $w(a)$, we formulate our purity-reweighted contrastive loss for all anchors $\mathcal{A}$ as follows,
\begin{equation}
\begin{aligned}
\mathcal{L}_\text{PC} = -\frac{1}{|\mathcal{A}|} \sum_{a \in \mathcal{A}} \log \frac{
    w(a) \cdot U^{+}(a)
  }{
    w(a) \cdot U^{+}(a) + U^{-}(a)
  },
\\
U^{+/-}(a) = \sum_{p \in P^{+/-}(a)} \exp ({\text{sim}(a, p)}/{\tau}), \label{eq:7}
\end{aligned}
\end{equation}
where \( P^+(a) = \{ p \in P \mid M_p = M_a \} \) and \( P^-(a) = \{ p \in P \mid M_p \neq M_a \} \) are positive set and negative set, respectively; \( \tau \) is a temperature parameter that controls smoothing; \( \text{sim}(a, p) \) is the similarity between \( a \) and \( p \).

\subsection{Total Loss}
During training, the proposed model is supervised by two loss functions, \textit{i.e.}, a standard cross-entropy loss \( \mathcal{L}_\text{CE} \) that serves to optimize segmentation results, and the proposed PC loss $\mathcal{L}_\text{PC}$ to optimize the prototype distributions. Specifically, the cross-entropy loss $\mathcal{L}_\text{CE}$ is defined as:
\begin{equation}
\mathcal{L}_\text{CE} = -\frac{1}{L} \sum \nolimits _{l=1}^{L} M_q^l \log(\hat{M}_q^l ), \label{eq:9}
\end{equation}
\noindent where 
\( M_q^l \)  and $\hat{M}_q^l$ represent GT mask and the predicted mask for query sample.

The overall loss is a weighted combination of $\mathcal{L}_\text{CE}$ and $\mathcal{L}_\text{PC}$ with a balancing weight $\lambda$, represented as:
\begin{equation}
\mathcal{L} _\text{total} = \mathcal{L}_\text{CE} + \lambda \cdot \mathcal{L}_\text{PC}. \label{eq:10}
\end{equation}

\begin{table*}[t]
\centering
\renewcommand{\arraystretch}{1.3} 
\setlength{\tabcolsep}{6pt} 
\begin{tabular}{@{\hspace{6pt}}l|@{\hspace{6pt}}ccc@{\hspace{22pt}}ccc@{\hspace{22pt}}ccc@{\hspace{22pt}}ccc@{\hspace{6pt}}}
\toprule
\multirow{2}{*}{Methods} & \multicolumn{3}{c@{\hspace{22pt}}}{1-way 1-shot} & \multicolumn{3}{c@{\hspace{22pt}}}{1-way 5-shot} & \multicolumn{3}{c@{\hspace{22pt}}}{2-way 1-shot} & \multicolumn{3}{c}{2-way 5-shot} \\ 
\cline{2-13}
 & $S^0$ & $S^1$ & mean & $S^0$ & $S^1$ & mean & $S^0$ & $S^1$ & mean & $S^0$ & $S^1$ & mean \\ 
\hline
AttMPTI & 36.32 & 38.36 & 37.34 & 46.71 & 42.70 & 44.71 & 31.09 & 29.62 & 30.36 & 39.53 & 32.62 & 36.08 \\
QGE & 41.69 & 39.09 & 40.39 & 50.59 & 46.41 & 48.50 & 33.45 & 30.95 & 32.20 & 40.53 & 36.13 & 38.33 \\
QGPA & 35.50 & 35.83 & 35.67 & 38.07 & 39.70 & 38.89 & 25.52 & 26.26 & 25.89 & 30.22 & 32.41 & 31.31 \\
COSeg  & 45.93 & 47.48 & 46.71 & 48.47 & 48.72 & 48.60 & 37.17 & 37.03 & 37.10 & 42.27 & 38.38 & 40.33 \\ 
\textbf{QHP (ours)} & \textbf{50.33} & \textbf{48.73} & \textbf{49.53} & \textbf{52.27} & \textbf{49.64} & \textbf{50.96} & \textbf{38.86} & \textbf{37.84} & \textbf{38.35} & \textbf{43.90} & \textbf{40.04} & \textbf{41.97} \\ 
\toprule
\end{tabular}
\caption{Comparisons of mIoU (\%) performance between our method and previous FS-3DSeg approaches on the \textbf{S3DIS} dataset. The best results are shown in \textbf{bold}.}
\label{table1}
\end{table*}

\begin{table*}[t]
\centering
\renewcommand{\arraystretch}{1.3} 
\setlength{\tabcolsep}{6pt} 
\begin{tabular}{@{\hspace{6pt}}l|@{\hspace{6pt}}ccc@{\hspace{22pt}}ccc@{\hspace{22pt}}ccc@{\hspace{22pt}}ccc@{\hspace{6pt}}}
\toprule
\multirow{2}{*}{Methods} & \multicolumn{3}{c@{\hspace{22pt}}}{1-way 1-shot} & \multicolumn{3}{c@{\hspace{22pt}}}{1-way 5-shot} & \multicolumn{3}{c@{\hspace{22pt}}}{2-way 1-shot} & \multicolumn{3}{c}{2-way 5-shot} \\ 
\cline{2-13}
 & $S^0$ & $S^1$ & mean & $S^0$ & $S^1$ & mean & $S^0$ & $S^1$ & mean & $S^0$ & $S^1$ & mean \\ 
\hline
AttMPTI & 34.03 & 30.97 & 32.50 & 39.09 & 37.15 & 38.12 & 25.99 & 23.88 & 24.94 & 30.41 & 27.35 & 28.88 \\
QGE & 37.38 & 33.02 & 35.20 & 45.08 & 41.89 & 43.49 & 26.85 & 25.17 & 26.01 & 28.35 & 31.49 & 29.92 \\
QGPA & 34.57 & 33.37 & 33.97 & 41.22 & 38.65 & 39.94 & 21.86 & 21.47 & 21.67 & 30.67 & 27.69 & 29.18 \\
COSeg  & 40.57 & 41.94 & 41.26 & 49.43 & 43.57 & 46.50 & 28.06 & 28.92 & 28.49 & 35.49 & 33.39 & 34.49 \\ 
\textbf{QHP (ours)}  & \textbf{40.70} & \textbf{42.92} & \textbf{41.81} & \textbf{50.10} & \textbf{44.80} & \textbf{47.45} & \textbf{28.45} & \textbf{29.07} & \textbf{28.76} & \textbf{36.11} & \textbf{34.30} & \textbf{35.21} \\ 
\toprule
\end{tabular}
\caption{Comparisons of mIoU (\%) performance between our method and previous FS-3DSeg approaches on the \textbf{ScanNet} dataset. The best results are shown in \textbf{bold}.}
\label{table2}
\end{table*}

\begin{figure*}[htbp]
\centering
\includegraphics[width=0.98\textwidth]{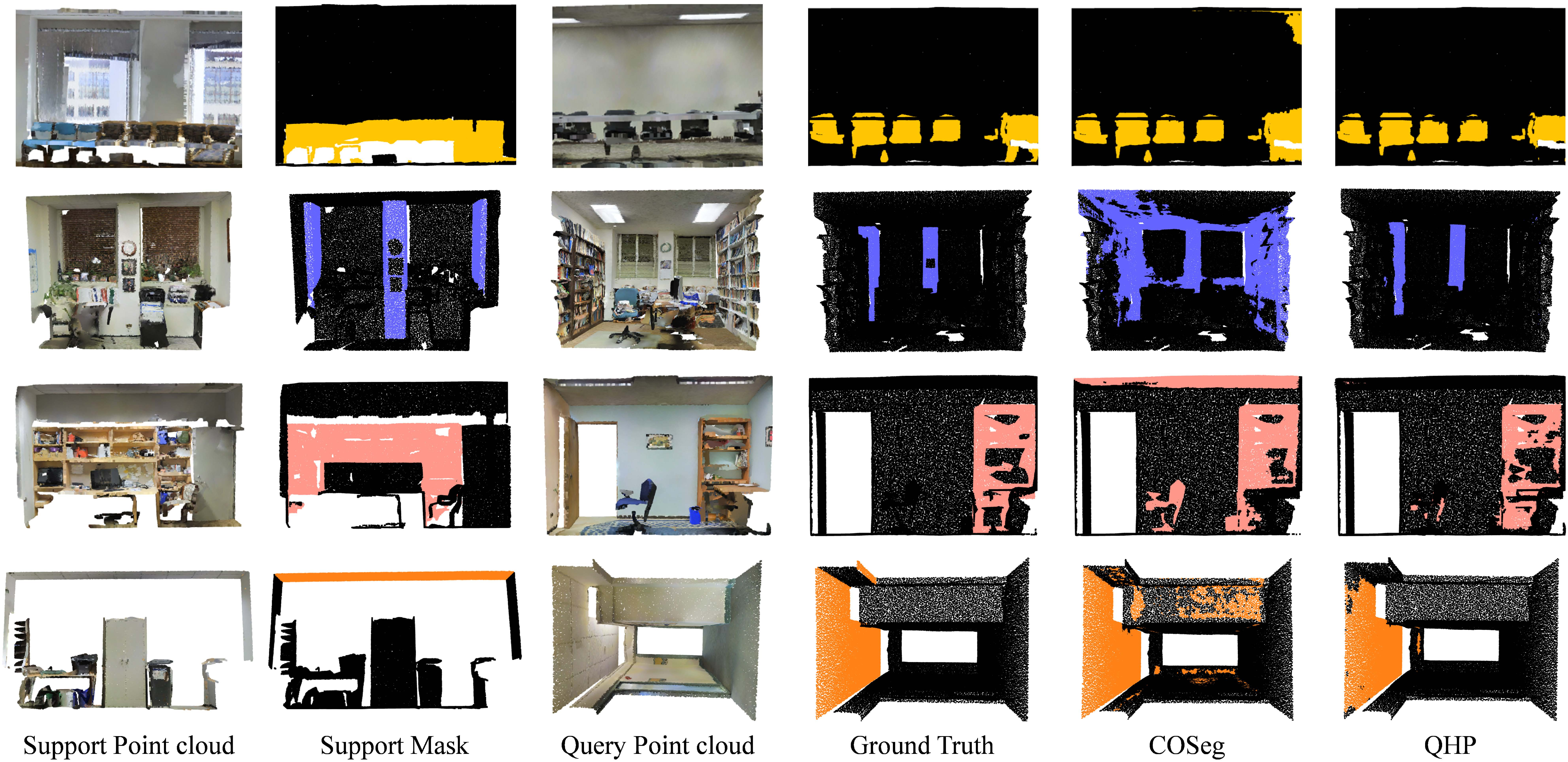} 
\caption{Qualitative comparisons between our proposed model QHP and COSeg. Each row, from top to bottom, represents the 1-way 1-shot task with the target category as chair (yellow), column (blue), bookcase (pink), and ceiling (orange), respectively.}
\label{fig4}
\end{figure*}

\section{Experiments}

\subsection{Datasets and Implementation Details}

\textbf{Datasets.} We conducted experiments on the S3DIS and ScanNet datasets. The S3DIS dataset \cite{s3dis} comprises five large-scale indoor areas across three buildings, annotated with 12 semantic classes for segmentation tasks. The ScanNet dataset \cite{ScanNet} contains 1,513 point cloud scans from 707 indoor scenes, covering 20 semantic categories. Compared to S3DIS, ScanNet features more irregular point clouds, rendering segmentation more challenging.
Data preprocessing follows \cite{an2024COSeg}: each room is divided into 1m × 1m blocks, input points are grid-sampled at 0.02m intervals, and after voxelization, 20,480 points are selected to standardize the input size.




\noindent \textbf{Training.}
We adopt data augmentation and pre-training protocols following COSeg \cite{an2024COSeg}, with the backbone pre-trained on each fold for 100 epochs. Meta-training is performed over 40,000 episodes using the AdamW optimizer, with a learning rate of \(5 \times 10^{-5}\) and weight decay of 0.01. For testing, we sample 1,000 episodes per class in 1-way settings and 100 episodes per combination in 2-way settings to ensure stable evaluations. We use 100 prototypes per class (\(\eta = 100\)). In \(k\)-shot scenarios where \(k > 1\), \(\eta/k\) prototypes are selected from each shot and concatenated to form the final prototypes. All models are implemented in PyTorch and trained on four NVIDIA RTX A4000 GPUs.


\noindent \textbf{Parameters.}
In both the HPG and PDO modules, the number of neighbors for hub point mining is set to \(k=5\). In the HPG module, the number of hub points/prototypes per class is set to \(\eta=100\). In the PDO module, the bad hub purity threshold is set to \(\gamma=0.6\) in  (Eq.~\ref{eq:6}). In the total loss function (Eq.~\ref{eq:10}), the balance weight for \(\mathcal{L}_\text{PC}\) is set to \(\lambda=0.1\).



\subsection{Comparison Results}
\noindent \textbf{Comparison with Previous Methods.} 
We compare our QHP with prior works including AttMPTI \cite{CVPR2021MPTI}, QGE \cite{ning2023QGE}, QGPA \cite{TIP2023QGPA}, and COSeg \cite{an2024COSeg} on S3DIS and ScanNet datasets.

$\bullet$~\textbf{S3DIS.} Table~\ref{table1} shows that our QHP consistently outperforms prior approaches across all settings. Specifically, compared to the baseline method COSeg, QHP achieves performance gains of \(2.82\%\) and \(2.36\%\) in the 1-way 1-shot and 1-way 5-shot settings, and enhancements of \(1.25\%\) and \(1.64\%\) in the 2-way settings. These gains can be attributed to the hub prototypes generated by our method. Unlike COSeg, which relies on FPS-based prototype generation that may produce redundant or irrelevant prototypes, our HPG module effectively identifies hub points to generate query-relevant prototypes, leading to more accurate segmentation outcomes. Additionally, optimizing prototype distributions further contributes to the superior performance of our model. When compared to query-guided methods such as QGE and QGPA, QHP demonstrates more significant advantages in the 1-way tasks, with improvements of \(9.14\%\) and \(2.46\%\), respectively. This highlights the superiority of our method in enhancing the discriminability of prototypes.

$\bullet$~\textbf{ScanNet.} Table~\ref{table2} shows that QHP outperforms all prior methods across all settings, further validating the effectiveness and applicability of our approach. Notably, in the 1-way 5-shot task, QHP achieves a mIoU of \(47.45\%\) and outperforms COSeg by \(1.19\%\), which highlights QHP's adaptability to the complex and diverse ScanNet dataset. We note that our performance gains are more pronounced in 5-shot than 1-shot settings: with more support samples available, QHP can mine important hub points from a larger pool of support points to generate query-relevant prototypes, thus capturing class-specific features more effectively. However, improvements on ScanNet are less substantial than S3DIS, as the dataset's higher complexity and inter-class overlap pose greater challenges for distinguishing similar categories.

\noindent \textbf{Qualitative Results.}
In Figure \ref{fig4}, we compare the results from our QHP (6th column) with those from COSeg (5th column). QHP improves object boundaries and category shapes, especially for column contours (blue, 2nd row), capturing finer details and reducing redundancy. The PDO module excels in the chair class (yellow, 1st row), resolving boundary ambiguities for more precise segmentation. Overall, QHP delivers cleaner, more accurate results with improved boundary delineation and reduced redundancy.

\begin{table}[t]
    \centering
    \setlength{\tabcolsep}{17pt} 
    \resizebox{0.43\textwidth}{!}{
    \begin{tabular}{@{\hspace{18pt}}ccc|c@{\hspace{18pt}}} 
        \toprule
        \textbf{Baseline} &  \textbf{HPG} & \textbf{PDO} & \textbf{mIoU (\%)} \\ 
        \midrule
        \checkmark &  &   & 45.93 \\ 
        \checkmark & \checkmark & & 47.37 \\ 
        \checkmark & \checkmark & \checkmark & \textbf{50.33} \\ 
        \bottomrule
    \end{tabular}
    }
    \caption{Ablation study of key components in QHP method.}
    \label{table3}
\end{table}

\subsection{Ablation Study}
We present an ablation study on the S3DIS dataset under 1-way 1-shot $S^0$ setting to validate the effectiveness of HPG and PDO modules, as well as hyperparameter settings.

\begin{figure}[htbp]
    \centering
    \begin{subfigure}[b]{0.45\columnwidth}  
        \centering
        \includegraphics[width=\linewidth]{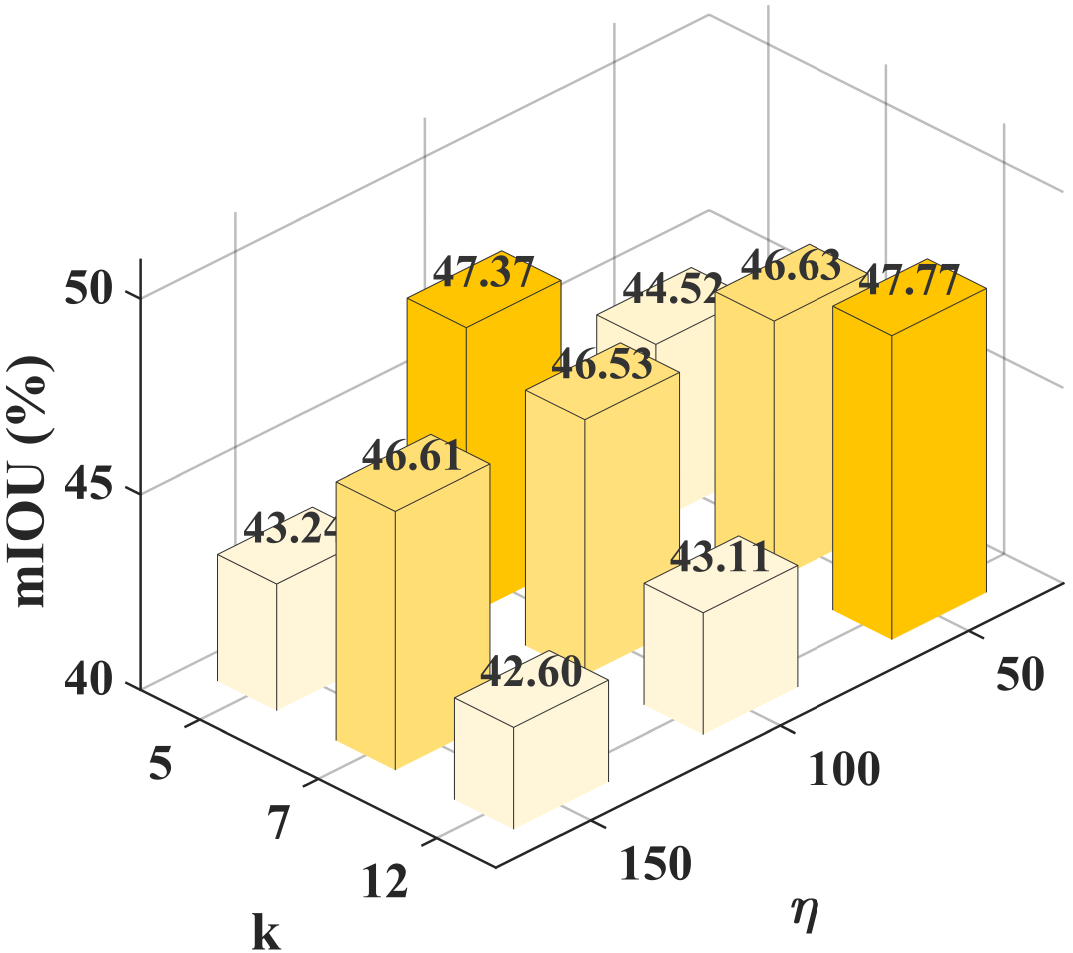}  
        \caption{$k$ and $\eta$ in HPG}  
        \label{subfig61}
    \end{subfigure}
    \hfill  
    \begin{subfigure}[b]{0.45\columnwidth}
        \centering
        \includegraphics[width=\linewidth]{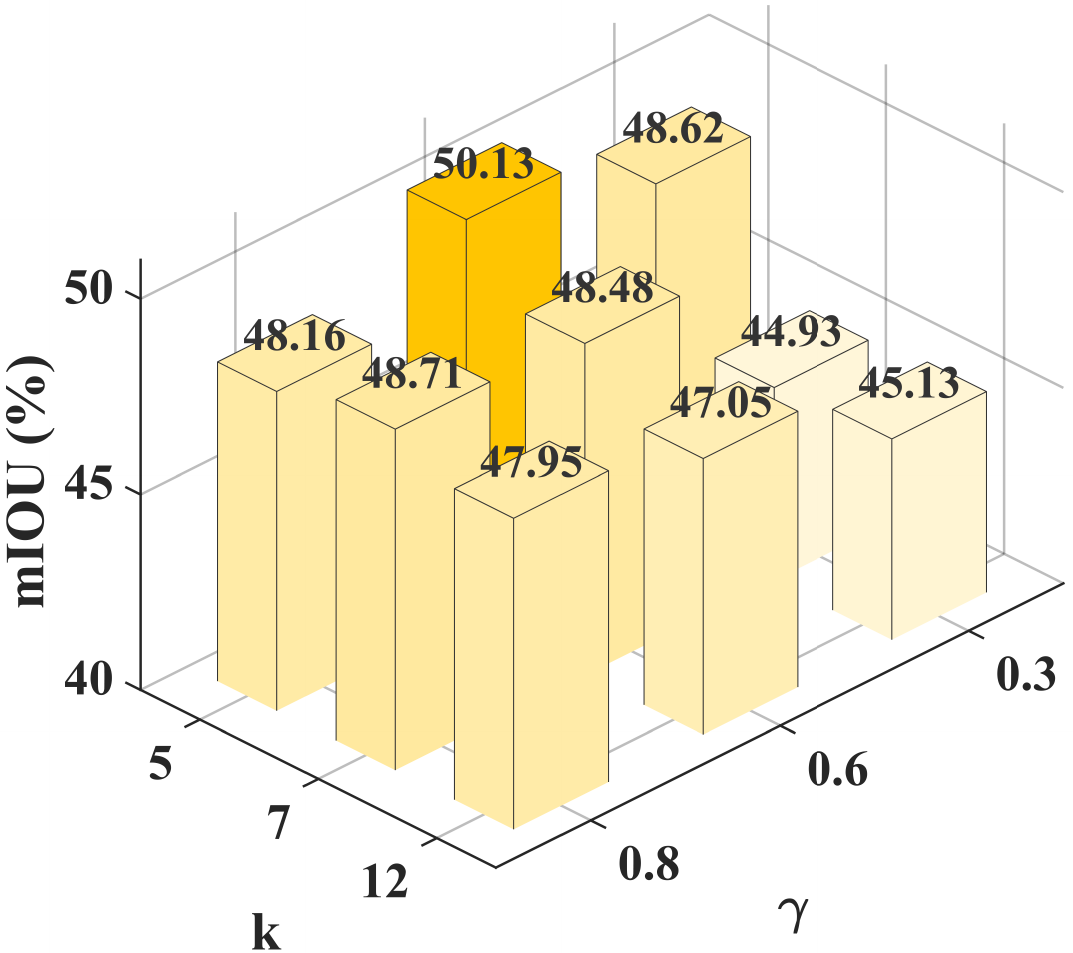}
        \caption{$k$ and $\gamma$ in PDO}
        \label{subfig62}
    \end{subfigure}
    \caption{Parameter sensitivity analysis of HPG and PDO.}
    \label{fig6}
\end{figure}

\noindent \textbf{Effects of Core Components.}
Using COSeg as the baseline, we conduct experiments to evaluate the effectiveness of the two core components, \textit{i.e.}, the HPG and PDO modules. As shown in Table~\ref{table3}, incorporating the HPG module improves the mIoU from \(45.93\%\) to \(47.37\%\) (\(+1.44\%\)), while the addition of the PDO module further increases the mIoU to \(50.33\%\) (\(+2.96\%\)), showing that the joint use of HPG and PDO significantly enhances the query relevance and discriminability of the prototypes, thereby leading to substantial performance gains.

\noindent \textbf{Effects of Hub Prototypes from HPG.}  
Without using the PDO module, we fix the number of prototypes to 100 per class and mix FPS-based prototypes with our hub prototypes in varying ratios.
As the hub prototype ratio increases from \(0\%\) (\textit{i.e.}, baseline COSeg) to \(50\%\), performance slightly drops, likely because prototype class diversity is reduced while the query relevance of prototypes remains limited, causing suboptimal performance. Beyond \(50\%\), performance steadily improves, peaking at \(100\%\), demonstrating that hub prototypes better capture support-query semantic correlation and provide more discriminative representations, significantly enhancing segmentation.

\begin{table}[t]
    \centering
    \resizebox{0.48\textwidth}{!}{
    \begin{tabular}{c|ccccc} 
        \toprule
        \textbf{Hub Prototype Ratio} & \textbf{0\%} & \textbf{30\%} & \textbf{50\%} &  \textbf{80\%} & \textbf{100\%}  \\ \midrule
        mIoU (\%)  & 45.93 & 42.93 &  42.32 & 45.47  & \textbf{47.37} \\ 
        \bottomrule
    \end{tabular}
    }
    \caption{Analyze the ratio of hub prototypes in HPG.}
    \label{table4}
\end{table}

\noindent \textbf{Impact of Parameters $k$ and $\eta$ in HPG.}
We analyze the impact of the number of neighborhoods $k$ in Eq.~\ref{eq:1} and the number of hub points $\eta$ in Eq.~\ref{eq:2}, as shown in Figure~\ref{fig6}(a). The best performance is achieved when $k = 12$ and $\eta = 50$, followed by $k = 5$ and $\eta = 100$. For fair comparison with other multi-prototype methods with 100 prototypes, we select the setting $k = 5$ and $\eta = 100$.

\noindent \textbf{Impact of Parameters \( k \) and \( \gamma \) in PDO.} 
We analyze the impact of the number of neighborhood \( k \) and purity threshold \( \gamma \) in PDO, as shown in Figure~\ref{fig6}(b). The best performance is achieved when \( k = 5 \) and \( \gamma = 0.6 \); thus, these parameters are selected in our setup.


\noindent \textbf{Effects of Different Contrastive Losses in PDO.}  
To verify the superiority of our proposed PC loss, we compared the effects of using the PC loss and the standard contrastive loss in the PDO module, as shown in Table~\ref{table7}. Our PC loss yields a \(0.51\%\) performance improvement, demonstrating that with the reweighting factor in our PC loss, both outlier prototypes and bad hubs near cluster boundaries are more effectively pulled into their respective class centers. This reduces boundary ambiguity and thus enhances overall performance.

\noindent \textbf{Impact of Coefficient \( \lambda \) in Total Loss.} 
Table~\ref{table8} illustrates the impact of the weight \( \lambda \) of the PC loss in Eq.~\ref{eq:10}. 
An appropriate value of \( \lambda = 0.1 \) yields the best results, indicating that the model achieves a balance between class boundary distribution and query segmentation. However, continued increases in \( \lambda \) could pull prototypes and hub points too tightly into clusters, thereby harming model performance.

We also provide more ablation studies in the \textit{Supplemental Material} to present more detailed designs of our model.

\begin{table}[t]
    \centering
    \setlength{\tabcolsep}{9pt} 
    \resizebox{0.48\textwidth}{!}{
    \begin{tabular}{@{\hspace{9pt}}c|cc@{\hspace{9pt}}}
        \toprule
        \textbf{Different Losses} & \textbf{Contrastive Loss} & \textbf{Our PC Loss} \\ \midrule
        mIoU (\%)  & 49.82 &  \textbf{50.33}\\ 
        \bottomrule
    \end{tabular}
    }
    \caption{Comparison between PC loss and contrastive loss.}
    \label{table7}
\end{table}

\begin{table}[t]
    \centering
    \resizebox{0.48\textwidth}{!}{
    \begin{tabular}{c|cccccc}
        \toprule
        \textbf{$\lambda$} & \textbf{0.9} & \textbf{0.7} &  \textbf{0.5} & \textbf{0.3} & \textbf{0.1} & \textbf{0} \\ 
        \midrule
        mIoU (\%)  & 38.28 &  44.06 & 45.08 & 44.10 & \textbf{50.33} &47.37\\ 
        \bottomrule
    \end{tabular}
    }
    \caption{Impact of the coefficient $\lambda$ for \(\mathcal{L}_\text{PC}\) in the total loss.}
    \label{table8}
\end{table}

\section{Conclusion}
We propose a Query-aware Hub Prototype (QHP) framework for few-shot 3D point cloud semantic segmentation, addressing limitations of prior methods relying solely on support prototypes. QHP models semantic correlations between support and query sets to enhance prototype relevance, with two key modules: a Hub Prototype Generation (HPG) module, identifying high-frequency hub points from support to generate query-relevant prototypes; and a Prototype Distribution Optimization (PDO) module, reducing the impact of bad hubs and ambiguous prototypes via purity-reweighted contrastive loss. Experiments on S3DIS and ScanNet show the superiority of the proposed model.
\bibliography{aaai2026}

\end{document}